\documentclass[runningheads]{llncs}
\usepackage{graphicx}
\usepackage{times}
\usepackage{amssymb}
\usepackage{soul}
\usepackage{multirow}
\usepackage{url}
\usepackage[utf8]{inputenc}
\usepackage[small]{caption}
\usepackage{graphicx}
\usepackage{amsmath}
\usepackage{booktabs}
\usepackage{breakcites}
\begin{document}
\title{Dynamic Joint Variational Graph Autoencoders}
%
\author{Sedigheh Mahdavi \and
Shima Khoshraftar\and
Aijun An}
\authorrunning{S. Mahdavi et al.}
%
\institute{Department of Electrical Engineering and Computer Science, York University \and
\email{\{smahdavi,khoshraf,aan\}@eecs.yorku.ca}}
\maketitle              
\begin{abstract}
Learning network representations is a fundamental task for many graph applications such as link prediction, node classification, graph clustering, and graph visualization. Many real-world networks are interpreted as dynamic networks and evolve over time. Most existing graph embedding algorithms were developed for static graphs mainly and cannot capture the evolution of a large dynamic network. In this paper, we propose Dynamic joint Variational Graph Autoencoders (Dyn-VGAE) that can learn both local structures and temporal evolutionary patterns in a dynamic network. Dyn-VGAE provides a joint learning framework for computing temporal representations of all graph snapshots simultaneously. Each auto-encoder embeds a graph snapshot based on its local structure and can also learn temporal dependencies by collaborating with other autoencoders. We conduct experimental studies on dynamic real-world graph datasets and the results demonstrate the effectiveness of the proposed method. 

\end{abstract}
\section{Introduction}

Many real world data can be formulated as graphs to represent complex relationships among the objects in the data. Dealing with high dimensional graph structures is a highly challenging task for many machine learning algorithms. Graph embedding methods are helpful to reduce the high dimensionality of graph data by learning low-dimensional features as latent representations. Many embedding algorithms \cite{tang2015line,grover2016node2vec,perozzi2014deepwalk,cao2015grarep,kipf2016variational,pan2018adversarially,goyal2018graph,hamilton2017inductive,cai2018comprehensive,cui2018survey,goyal2018graph,cao2015grarep,wang2017community,tu2016max,ou2016asymmetric,cao2016deep} have been proposed to capture different characteristics of a network and they provide effective ways to extract low-dimensional latent representations of graphs.

Most of the embedding approaches are designed as static methods, assuming that the nodes and edges in the graph are fixed. However, real networks are often dynamic, consisting of vertices and edges that may occur and disappear at different time points. In modeling a dynamic network, it is essential to take temporal dependencies and changes into account for characterizing evolving nodes and edges. Capturing these temporal factors requires a dynamic model that can learn the evolution track of a dynamic network over time. A dynamic network is often represented as a sequence of static graph snapshots over time. Some attempts have been made to develop a dynamic model for leaning temporal low-dimensional latent representations of graph snapshots \cite{goyal2018dyngem,zhuscalable,goyal2018capturing,zhou2018dynamic,mahdavi2018dynnode2vec,nguyen2018continuous,trivedi2018representation,yu2018netwalk}. A simple traditional method for obtaining the embedding vectors of a graph snapshot is computing embedding vectors in each timestamp separately. Then, the graph embedding vectors across all timestamps are aligned and placed in a same vector space.

Alignment methods raise two main issues. First, some learned embedding vectors are invariant to transformations and it is not possible to place embedding vectors of all graph snapshots in the same latent space. Second, such methods need to solve a separate optimization problem for finding transformation functions. In \cite{zhuscalable}, the authors proposed a joint matrix factorization-based optimization function which can jointly find embedding vectors across time without the alignment step. The problem of this joint optimization function is a non-convex and non-linear optimization problem with a large number of variables. Recently, different types of deep autoencoders have been proposed to solve complex nonlinear functions \cite{pan2018adversarially,hamilton2017inductive,wang2016structural}. Among the autoencoders, variational graph auto-encoder (VGAE) \cite{kipf2016variational} is an effective static embedding method which is a deep generative model by variational inference. Similar to other static graph embedding methods, it is not designed for the dynamic setting. 
Joint deep learning models have recently achieved great success in many learning tasks such as multi-task learning and domain adaptation \cite{epstein2018joint}, multimodal network embedding \cite{xu2017embedding}, video description generation \cite{pan2016jointly}, image-text representation \cite{huang2018multimodal}, and clustering \cite{ghasedi2017deep}. These joint deep leaning models aim to learn dependencies and similarities among some target tasks by joining their goals together. As a dynamic network includes temporal dependencies among all graph snapshots, the intuition of join deep learning motivates us to develop a joint autoencoder based on VGAE.

In this paper, we propose a joint dynamic Variational autoencoder (Dyn-VGAE) which simultaneously learns latent representations of the graphs at all timestamps. To the best of our knowledge, this is the first study on developing a deep joint learning for dynamic network representations. We first assign a specific variational autoencoder with a modified learning function for each graph snapshot. This learning framework jointly learns latent variables of each graph and captures evolving patterns among graphs by sharing learned latent variables during the training iterations. The main contributions of this work are:
\begin{itemize}
    \item We introduce a joint learning approach, which is the first study to extract the dynamic network representation with collaboration among graph autoencoders. During the training steps, each autoencoder can collaborate with other autoencoders of previous graph snapshots. 
    \item We define a novel probabilistic smoothness term in the loss function to align latent spaces across time, which provides a transfer learning strategy to adjust learned latent spaces.  
    \item We conducted experiments on dynamic real-world datasets, which show that the proposed joint method can significantly improve the state of the art.
\end{itemize}
The rest of the paper is organized as follows. The related works are discussed in Section 2. In Section 3, we describe our proposed architecture. The experiments are detailed in Section 4. Conclusions and future work are presented in Section 5.

\section{Related Work}
In this section, we describe related work on static, dynamic, and joint deep learning methods.   
\subsection{Static graph embeddings}
Network embedding for static graphs has been well studied in recent years \cite{tang2015line,grover2016node2vec,perozzi2014deepwalk,cao2015grarep,kipf2016variational,pan2018adversarially,goyal2018graph,cai2018comprehensive,hamilton2017inductive,hamilton2017representation}. Conventional static network embedding methods include DeepWalk\cite{perozzi2014deepwalk}, Node2vec\cite{grover2016node2vec}, LINE \cite{tang2015line} and SDNE\cite{wang2016structural}. \cite{perozzi2014deepwalk,grover2016node2vec,tang2015line} are based on random walks and utilize random walks for capturing the neighborhood of the nodes in a graph inspired by advances in natural language processing. The main difference among these methods is related to the special kind of random walk they exploit. In \cite{perozzi2014deepwalk} and \cite{grover2016node2vec}, uniform random walks and BFS/DFS-like random walks are used, respectively. Another type of graph embedding methods \cite{kipf2016variational,pan2018adversarially,wang2016structural,hamilton2017inductive} is based on adjacency matrices of graphs. In \cite{wang2016structural}, the first-order and second-order proximity is used to represent the network structure. In \cite{pan2018adversarially} a jointly optimized adversarial framework is proposed for network embeddings. In this framework, the method first learns the topology of the graph and then forces the latent codes to be similar to a prior distribution.

\subsection{Dynamic graph embeddings}
Temporal network representation is a challenging problem as it needs to consider the evolutionary structure of graphs over time. There have been some recent efforts to tackle the problems in dynamic network embedding \cite{goyal2018dyngem,zhuscalable,goyal2018capturing,zhou2018dynamic,mahdavi2018dynnode2vec,nguyen2018continuous,trivedi2018representation,goyal2018dynamicgem,kumar2018learning}. \cite{zhou2018dynamic} uses the triadic closure process to learns changes in the structure of graphs. This method is specifically designed for undirected graphs. \cite{goyal2018dyngem} is a deep learning model that initialized a graph model with weights from models of previous graphs to create the desired alignment among temporal embeddings. In \cite{zhuscalable} a joint matrix factorization method was proposed that learns a temporal latent space model for dynamic networks by developing local and incremental block-coordinate gradient descent algorithms. In \cite{goyal2018capturing}, the authors proposed a deep learning method that captures the transition model of dynamic networks using dense and recurrent layers. dyngraph2vecAE, dyngraph2vecRNN and dyngraph2vecAERNN are three variations in \cite{goyal2018capturing}. dyngraph2vecAE models the changes in graphs using multiple fully connected layers. Then, dyngraph2vecRNN is presented as it has less parameters and takes into consideration the long term dependencies in graphs using  an LSTM structure. In dyngraph2vecAERNN, the dimension of input vectors to LSTM is reduced by inputting node representation vectors into LSTM rather than sparsed vectors.

\subsection{Joint deep learning methods}

Recently, many algorithms have been proposed to jointly learn embeddings for different applications. In \cite{epstein2018joint}, a joint learning framework was developed for multiple tasks. The authors designed a neural network model with shared branches for extracting information of common features and local branches for learning features of each task. For visual semantic embeddings, Pan et al. \cite{pan2016jointly} introduced a long short-term memory with a visual-semantic embedding architecture which simultaneously learns the semantic sentence and video content. A joint convolutional autoencoder was proposed in \cite{ghasedi2017deep} for the clustering task by jointing clustering and embedding tasks. Ren et al. \cite{ren2016joint} introduced a joint representation for image-text embedding task using the visual information in the text model. A joint embedding was introduced in \cite{xu2017embedding} for coupled networks. Each network transfers some relevant information to other networks for learning intra-network edges in these networks. Huang et al. \cite{huang2018multimodal} proposed a deep joint embedding which incorporates the link information and multimodal contents together to obtain embedding for social media.

\section{Method}

A dynamic network is represented as a time-ordered sequence of static graphs, $G_1,G_2,$
$\dots,G_T$, where $T$ is the number of time steps. The graph at time $t$ is denoted by $G_t=(V_t,E_t)$ with a set $V_t$ of $|V|$ vertices and an edge set $E_t$ that may change in the time interval $[0, T]$. The dynamic graph embedding can be formulated as a temporal mapping function $ f_t: A_t \rightarrow{Z_t}$ which finds a low-dimensional latent representation $Z_t$ for graph $G_t$ with an adjacency matrix $A_t$.  In this section, we first review the static variational graph autoencoder briefly and then propose a novel dynamic graph embedding method, which we call Dynamic joint Variational Graph Autoencoders (Dyn-VGAE).

\subsection{Static variational graph autoencoder (SVGAE)}

The overall architecture of SVGAE consists of two components: a variational graph convolutional network encoder and a probabilistic decoder \cite{zhou2018dynamic}. The variational graph encoder is defined by an inference model, which encodes the observed graph data into stochastic low-dimensional latent variables ($Z$). The variational graph decoder is designed by a generative model, which decodes latent variables into the distribution of the observed graph data. Let $G = (V, E)$ denote a graph with an adjacency matrix $A$ and the content features $X$, where $V$ and $E$ are nodes and edges of the graph. 
Variational graph convolutional encoder is constructed as follows:
\begin{align*}
GCN(X, A) &= (D^{-1/2}AD^{-1/2}) f_{R} ((D^{-1/2}AD^{-1/2})XW_0)W_1 \\
f_{R} (t)&=Relu(t)= max(0, t) 
\end{align*}
where $D$ is a degree matrix. Weight matrices are $W_i$ and first-layer parameters $W_0$ are shared between $GCN_\mu(X, A)$ and $GCN_\sigma(X, A)$. The generative process is characterized by an inner product between latent variables:
\begin{align*}
p (A|Z) &= \prod_{i=1}^{N}\prod_{j=1}^{N} p(A_{i,j}| z_i, z_j ), \\
with \hspace{0.1cm} p (A_{i,j} = 1| z_i, z_j ) &= sigmoid(z_i^T,z_j )
\end{align*}
The inference process is modeled by a two-layer graph convolutional network (GCN) \cite{kipf2016semi} (Variational Graph Encoder) :
\begin{align*}
 q(Z|X, A)= \prod_{i}^{N} q(z_i|X, A), with \hspace{0.1cm}q(z_i| X, A)= \mathcal{N} (z_i |\mu_i,diag(\sigma^2_i ))
\end{align*}
where $N = |V|$, $\mu$ and $\sigma$ are parameters of the Gaussian distribution $q(.)$, $\mu = GCN_\mu(X, A)$ is the matrix of mean vectors, and $log\sigma = GCN_\sigma(X, A)$.
The variational autoencoder is trained by maximizing the variational lower bound $L_{VLBO}=E_{q(Z|X,A)}[log p (A|Z)] -KL[q(Z | X, A)|| p(Z)]$, where $KL[p(.)||q(.)]$ is the Kullback-Leibler (KL) divergence between $p(.)$ and $q(.)$. The $L_{VLBO}$ is usually optimized via stochastic gradient descent, using the reparameterization trick to estimate the gradient. 

\subsection{Dynamic joint variational graph autoencoders (Dyn-VGAE)}
Let $G_1,G_2,\dots,G_T$ denote a dynamic network with a series of adjacency matrices $A_1,A_2,\dots,A_T$. Dyn-VGAE aims to obtain a low dimensional latent representation of each graph $G_t$. This representation preserves both the local topology and the structure of a static graph snapshot $G_t$ and also captures its evolutionary pattern from the previous time steps. In the proposed joint framework, each graph $G_t$ has its own model (the variational autoencoder $VGAE_t$) which is similar to SVGAE except that it has a different learning loss function. The joint learning function encourages all autoencoders to collaborate together for obtaining similar parameters (latent representations). We describe the algorithm in detail below.
\subsubsection{Autoencoder model for graph $G_t$}
The encoder of a graph snapshot $G_t$ with an adjacency matrix $A_t$ and the content features $X _t$ is modeled by a two-layer GCN:
\begin{align*}
q_t(Z_t|X_t, A_t)&= \prod_{i}^{N_t} q_t(z_{i}^t|X_t, A_t),\\
with \hspace{0.1cm} q_t(z_{i}^t| X_t, A_t)&= \mathcal{N} (z_{i}^t |\mu_{t,i},diag(\sigma_{t,i}^2 ))
\end{align*}
The probabilistic decoder of $VGAE_t$ is:
\begin{align*}
p_t (A_t|Z_t) = \prod_{i=1}^{N_t}\prod_{j=1}^{N_t} p_t(A_{i,j}^t| z_{i}^t, z_{j}^t ), \\  with \hspace{0.2cm}
p_t (A_{i,j}^t = 1| z_{i}^t, z_{j}^t ) = sigmoid({(z_{i}^t)}^T,z_{j}^t ). 
\end{align*}
Similar to the static variational autoencoder, $VGAE_t$ optimizes the variational lower bound for learning the current latent representation by minimizing the loss function as follows:
\begin{align*}
 \mathit{min} \hspace{0.15cm}  L_{V_t} =E_{q_t(Z_t|X_t,A_t)}[log p_t (A _t| Z_t)]\\ -\mathcal{KL}[q_t(Z_t | X_t,A_t)|| p_t(Z_t)]
\end{align*}
A general assumption of a dynamic network is \cite{sarkar2006dynamic,goyal2018dyngem,mahdavi2018dynnode2vec,zhou2018dynamic} that changes are smooth and continuous in a short duration (length $l$) \cite{zhou2018dynamic}. 
Thus, the key question here is how an associated encoder for graph $G_t$ can learn aligned embedding vectors with embedding vectors of other graphs in a dynamic network. 
We change the learning process in which the autoencoder $VGAE_t$ can be joint with other autoencoders of previous graphs during the training process.

By collaborating with other prior autoencoders, each autoencoder is able to transfer temporal dependencies from previous latent spaces to the current latent space. The parameters of $VGAE_t$ are obtained by a modified loss function which has a temporal smoothness dependency term for aligning the latent space of graph $G_t$ with $l$ prior snapshots. By assuming change smoothness, we force the current latent representations to be similar to the previous latent vectors by minimizing the difference between two distributions of the current latent space and a temporal Gaussian random walk \cite{sarkar2006dynamic}. The temporal Gaussian random walk is defined based on latent representations of the previous graphs in $l$ prior times. For simplicity, first we explain our method by assuming the length $l$ is equal to two. This means the current latent space ($Z_t$) of the graph $G_t$ should be similar to only the previous latent space ($Z_{t-1}$); then we will extend $l$ to a general length. This temporal Gaussian random walk ($q_W^t$) can be defined as a Gaussian distribution with the mean $Z_{t-1}$:
\begin{align*}
q_W^t= \mathcal{N}(Z_{t-1},\sigma^2)
\end{align*}  
where $\sigma^2$ is considered as Gaussian noise with a fixed standard deviation \cite{sarkar2006dynamic}. Temporal smoothness dependency term $\mathcal{L}_{s}^t$ is defined as the Kullback-Leibler ($\mathcal{KL}$) divergence among $q_W^t$ and $q(Z_t|X_t, A_t)$ :
\begin{align*}
\mathcal{L}_s^t  =\mathcal{KL}[q_t(Z_t | X_t, A_t)|| q_W^t]
\end{align*}
$\mathcal{L}_{s}^t$ prevents the current latent vectors from being placed very far from latent vectors in the previous timestamps. Then, the final learning loss function can be formulated by combining the variational learning function $\mathcal{L}_v^t$ and temporal smoothness dependency term $\mathcal{L}_{s}^t$:
\begin{equation*}
\mathit{min}\hspace{0.1cm} \mathcal{L}_C^t = \mathcal{L}_v^t+\gamma\mathcal{L}_s^t 
\end{equation*}
where the hyperparameter $\gamma$ controls the importance of the two losses. The term $\mathcal{L}_v^t$ learns
latent representations of the graph $G_t$ by minimizing the distance between the model prediction $p(.)$ and the
target variable $q(.)$. The additional smoothness term $\mathcal{L}_s^t$ forces latent representations to be 
aligned with prior latent representations of the graph $G_t$.For $l>2$, $\mathcal{L}_v^t $ can be formulated as follows:
\begin{align*}
\mathcal{L}_{s}^t = \sum_{i=t}^{t-l}\mathcal{KL}[q_t(Z_t|X_t, A_t)||q_W^i]
\end{align*}
\begin{figure*}
    \centering
    \includegraphics[scale=0.6]{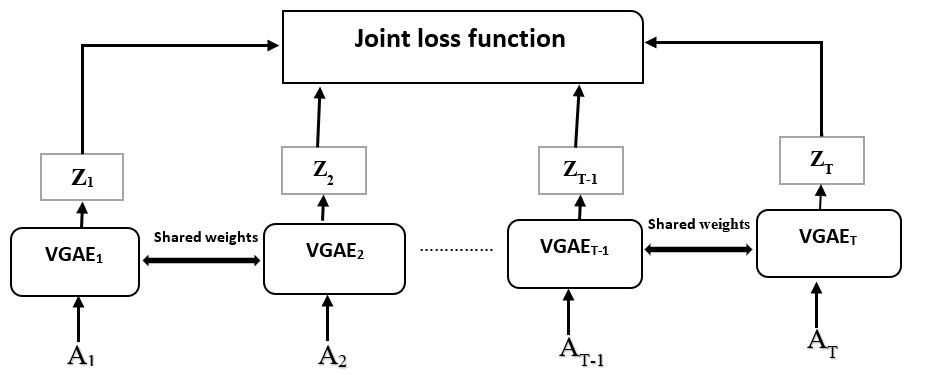}
    \caption{Joint framework for graphs $G_1,G_2,\dots,G_T$ with a series of adjacency matrices $A_1,A_2,\dots,A_T$}
    \label{fig1}
\end{figure*}

\subsubsection{Joint dynamic graph autoencoders framework}

The joint learning framework is shown in Figure \ref{fig1}. Consider $G_1,G_2,\dots,G_T$ as a dynamic network, we assign  $T$  autoencoders $VGAE_1,VGAE_2,\dots,VGAE_T$ for all graphs.  All loss learning functions of these autoencodres can be jointly collaborated while each autoencoder can focus on its own task to learn its own graph latent representations. The joint loss learning function can be formulated as the summation of the loss function of all autoencoders:
\begin{align*}
\mathit{min}\hspace{0.1cm} \sum_{i=1}^T \mathcal{L}_C^i =\sum_{i=1}^T \mathcal{L}_v^i+\gamma\mathcal{L}_s^i \\ 
=\sum_{i=1}^T [\prod_{j=1}^{N_i}\prod_{k=1}^{N_i} p_i(A_{j,k}^i| z_{j}^i, z_{k}^i)+\\
\sum_{j=i}^{i-l} \mathcal{KL}[q_i(Z_i | X_i, A_i)||q_W^j]]
\end{align*}
Each autoencoder is trained to learn latent variables specific to a graph and extract temporal dependencies among graphs by sharing learned latent variables during training iterations. Therefore, the embedding latent representations for each timstamp can be aligned jointly with $l$ autoencoders of previous graph snapshots by using shared weights.

We use the reparameterization trick mentioned in \cite{kingma2013auto,kipf2016variational} to minimize the loss function. During training steps, weights for an autoencoder jointed with $l$ other autoencoders can be updated in two strategies. In the first method, gradients in the autoencoder $t$ are computed with respect to fixed weights from the previous update step of other autoencoders. The second strategy updates gradients of the autoencoder $t$ after getting new updates of other autoencoders. In this paper, we choose the second strategy which provides a flexible framework to train autoencoders. Before the training step, we find the common nodes among each graph with its $l$ previous graphs, so during training each autoencoder just needs to share learned weights for common nodes. The proposed joint framework is very practical because the gradient of the additional smoothing term can be easily computed similar to the $KL$ term in the variational autoencoder. Also, the framework is trivially parallelizable and each autoencoder just needs to cooperate with only $l$ autoencoders where $l$ is considered as a short period of time.  

\section{Experiments}
We performed the evaluation of our method on multiple real-world datasets from various domains on node classification, link prediction and receommendation tasks. The findings of our experiments are reported as follows. 
\subsection{Baselines} 
The models for comparison are listed below:
\begin{itemize}
    \item \textbf{DeepWalk \cite{perozzi2014deepwalk}:} DeepWalk is a static network embedding method based on uniform random walks.
    \item \textbf{Node2vec \cite{grover2016node2vec}:} This method is a static network representation algorithm utilizing breadth-first-search (BFS) or depth-first-search (DFS) based random walks and skipgram.
    \item \textbf{SVGAE \cite{kipf2016variational}:} This is a variational graph autoencoder model that works for static graphs. SVGAE is an inference-based graph embedding model that encodes the observed graphs into their respective  distribution.
    \item \textbf{dynAE \cite{goyal2018dyngem}:} dynAE stands for dyngraph2vecAE, a dynamic network embedding method based on dyngraph2vec. It utilizes deep learning models with multiple fully-connected layers to model interconnections of nodes.
    \item \textbf{dynAERNN \cite{goyal2018dyngem}:} dynAERNN is the short form we used for dyngraph2vecAERNN. This is another variant of the dyngraph2vec method, which is a dynamic representation learning method. It feeds previously learned representations to LSTMs to generate embedding vectors.
\end{itemize}
For each static baseline method, we apply the static baseline method independently to each graph snapshot $G_t$ in a dynamic network.

\subsection{Experiment Settings}
We run our experiments on DeepWalk and Node2vec with $(p,q)=(1,1)$ and $(p,q)=(0.5,1)$, respectively. The number of random walks per node is set to 10. For dynAE and dynAERNN, we used the default parameters in the publicly available source code \cite{goyal2018dyngem,goyal2018dynamicgem}. The parameters of SVGAE and Dyn-VGAE are similar to \cite{kipf2016variational}. Their autoencoder model structure consists of 32 dimensional hidden layers and 16 dimensional latent variables. For training, we used the Adam optimizer, the learning rate is 0.01, and number of epochs are 200. Dyn-VGAE1 is Dyn-VGAE with $l=1$ and in Dyn-VGAE2, $l=2$.

\begin{table}[]
\begin{center}
\caption {Macro-F1 and Micro-F1 scores for node classification} \label{tab1:title}
\begin{tabular}{l|l|l|l|l}
\multicolumn{1}{c|}{\multirow{2}{*}{\textbf{Method}}} & \multicolumn{2}{c|}{\textbf{Acm}} & \multicolumn{2}{c}{\textbf{Dblp}} \\ \cline{2-5} 
\multicolumn{1}{c|}{} & mac-f1 & mic-f1 & mac-f1 & mic-f1 \\ \hline
node2vec & 0.3775 & 0.5221 & 0.3768 & 0.5185 \\
DeepWalk & 0.3532 & 0.502 & 0.3815 & 0.5245 \\
SVGAE & 0.3896 & 0.5664 & 0.4224 & 0.5227 \\
dynAE & 0.3699 & 0.5237 & 0.3675 & 0.479 \\
dynAERNN & 0.402 & 0.5581 & 0.3876 & 0.4959 \\ \hline
\textbf{Dyn-VGAE1} & 0.4048 & 0.575 & 0.439 & 0.5283 \\
\textbf{Dyn-VGAE2} & \textbf{0.4402} & \textbf{0.5896} & \textbf{0.4716} & \textbf{0.5356}
\end{tabular}
\end{center}
\end{table}

\begin{table}[]
\begin{center}
\caption {AUC scores for link prediction} \label{tab2:title}
\begin{tabular}{l|l|l|l}
\multicolumn{1}{c|}{\textbf{Method}} & \multicolumn{1}{c|}{\textbf{Hep-th}} & \multicolumn{1}{c|}{\textbf{AS}} & \multicolumn{1}{c}{\textbf{St-Ov}} \\ \hline
node2vec & 0.973137 & 0.91395 & 0.59249 \\
DeepWalk & 0.97238 & 0.91219 & 0.58776 \\
SVGAE & 0.97499 & 0.91974 & 0.65437 \\
dynAE & 0.87834 & 0.7969 & 0.52017 \\
dynAERNN & 0.93851 & 0.83913 & 0.56149 \\ \hline
\textbf{Dyn-VGAE1} & \textbf{0.98236} & 0.92981 & \textbf{0.74065} \\
\textbf{Dyn-VGAE2} & 0.97754 & \textbf{0.93187} & 0.69466
\end{tabular}
\end{center}
\end{table}

\subsection{Node Classification}
In node classification tasks, each node in a graph has a class label. We predict the class label for the nodes in graph $G_t$ using previous graphs in the stream from $0$ to $t-1$ based on the approach mentioned in \cite{goyal2018dyngem}. Our classification method is logistic regression. We used two measures, Micro-f1 and Macro-f1, for evaluating our method. The results are presented in Table \ref{tab1:title}. The datasets are as follows:
\begin{itemize}
    \item \textbf{Dblp \cite{tang2008arnetminer,tang2008extraction}:} Dblp is the main coauthorship network of researches in various fields with 90k nodes and 749k edges over 18 years (2000-2017). There are two class labels for nodes: 1) database and data mining (VLDB, SIGMOD, PODS, ICDE, EDBT, SIGKDD, ICDM, DASFAA, SSDBM, CIKM, PAKDD, PKDD, SDM and DEXA) 2) computer vision and pattern recognition (CVPR, ICCV, ICIP, ICPR, ECCV, ICME and ACM-MM).
    
    \item \textbf{Acm \cite{tang2008arnetminer,tang2008extraction}:} The Acm dataset has the same characteristics as the Dblp dataset. The timespan of Acm is considered as 16 years (2000-2015).
\end{itemize}
Based on the results, it is evident that our approach outperforms the baselines in both Acm and Dblp datasets. Specially, Macro-F1 scores are significantly better than the closest benchmarks and our performance gain is above other methods in terms of Micro-F1 scores. From the results, it can been seen that on both Acm and Dblp datasets, increasing the effect of previous graphs by extending $l$ in Dyn-VGAE2 improves the overall results. The reason is that in  coauthorship datasets the changes between consecutive snapshots are smooth and the research area of authors is fixed in a short period of time. 

\subsection{Link Prediction}
One of the main graph mining tasks is link prediction as it shows the effectiveness of the edge embeddings in predicting unseen edges. We predict edges in graph $G_t$ using previous learned embeddings of graph $G_{t-1}$ mentioned in \cite{goyal2018dyngem}. For this task, the reconstruction scores are computed similar to \cite{kipf2016variational} and we report the average AUC (area under ROC curve) scores over time from $1$ to $T$ for all datasets in Table \ref{tab2:title}. The evaluation was performed on the following three datasets.
\begin{itemize}
    \item \textbf{Hep-th \cite{leskovec2015snap}:} This is the coauthorship network of researchers in high energy physics theory conference with 34k nodes, 421k edges, 60 time points.
    \item \textbf{AS \cite{leskovec2015snap}:} Autonomous Systems are the communication network between users in BGP. It contains 6k nodes, 13k edges and 100 time steps.
    \item \textbf{St-Ov \cite{leskovec2015snap}:} This dataset shows the user interactions in the Math Overflow website. This dataset consists of 14k nodes and 195k edges over 58 time points.
\end{itemize}
The results show that Dyn-VGAE achieves the highest AUC in all the three datasets. In AS and St-Ov, our approach outperforms the benchmark methods by a significant margin. This highlights that our method effectively learns the dynamic representations in these datasets. Similarly, the results of our method on the Hep-th dataset are better than those of other methods. The effects of increasing $l$ for AS and Hep-th are not that significant while this is not the case in St-Ov. In the St-Ov dataset, graph snopshots have less common edges. Therefore, its results with smaller $l$'s are better.

\begin{table}[!]
\begin{center}
\caption {Analysis of parameter $\gamma$ for node classification} \label{tab3:title}
\begin{tabular}{c|l|r|l}
\multicolumn{1}{l|}{\textbf{Dataset}} & \textbf{$\gamma$} & \textbf{mac-f1} & \textbf{mic-f1} \\ \hline
\multirow{3}{*}{\textbf{Dblp}} & 0.5 & 0.4386 & 0.5288 \\
 & 1.5 & 0.4377 & 0.5299 \\
 & 2 & \multicolumn{1}{l|}{0.4362} & 0.529 \\ \hline
\multirow{3}{*}{\textbf{Acm}} & 0.2 & 0.3992 & 0.5714 \\
 & 1.2 & 0.4028 & 0.5754 \\
 & 1.5 & 0.4004 & 0.5776
\end{tabular}
\end{center}
\end{table}
\begin{table}[!]
\begin{center}
\caption {Analysis of parameter $\gamma$ for link prediction} \label{tab4:title}
\begin{tabular}{c|l|r}
\multicolumn{1}{l|}{\textbf{Dataset}} & \textbf{$\gamma$} & \multicolumn{1}{c}{\textbf{AUC}} \\ \hline
\multirow{3}{*}{\textbf{Hep-th}} & 0.7 & 0.98186 \\
 & 1 & 0.98031 \\
 & 1.2 & \multicolumn{1}{l}{0.97773} \\ \hline
\multirow{3}{*}{\textbf{AS}} & 0.2 & 0.92283 \\
 & 0.5 & 0.92858 \\
 & 1.2 & 0.92766 \\ \hline
\multirow{3}{*}{\textbf{St-Ov}} & 0.5 & \multicolumn{1}{l}{0.75342} \\
 & 0.7 & \multicolumn{1}{l}{0.72256} \\
 & 1 & \multicolumn{1}{l}{0.72064}
\end{tabular}
\end{center}
\end{table}

\begin{table*}[]
\caption{Precision and Recall for recommendation on AS dataset}
\label{tab5}
\small
\begin{tabular}{l|l|lllllllll}
\multicolumn{1}{c|}{\textbf{Method}} & \multicolumn{1}{c|}{} & \multicolumn{1}{c}{\textbf{k=2}} & \multicolumn{1}{c}{\textbf{k=3}} & \multicolumn{1}{c}{\textbf{k=4}} & \multicolumn{1}{c}{\textbf{k=5}} & \multicolumn{1}{c}{\textbf{k=6}} & \multicolumn{1}{c}{\textbf{k=7}} & \multicolumn{1}{c}{\textbf{k=8}} & \multicolumn{1}{c}{\textbf{k=9}} & \multicolumn{1}{c}{\textbf{k=10}} \\ \hline
\multirow{2}{*}{node2vec} & Recall & \multicolumn{1}{r}{0.214669} & 0.334204 & 0.367184 & 0.394506 & 0.407605 & 0.416419 & 0.425321 & \textbf{0.429238} & \textbf{0.4371} \\
 & Precision & \multicolumn{1}{r}{0.388217} & 0.543356 & 0.640374 & 0.696155 & 0.739523 & 0.759523 & 0.769879 & \textbf{0.781488} & \textbf{0.785693} \\ \cline{2-11} 
\multirow{2}{*}{DeepWalk} & Recall & \multicolumn{1}{r}{0.220448} & 0.338428 & 0.368539 & 0.394732 & 0.407542 & 0.415795 & 0.423611 & 0.42708 & 0.434744 \\
 & Precision & \multicolumn{1}{r}{0.395229} & 0.547698 & 0.64106 & 0.695375 & 0.737814 & 0.756948 & 0.76571 & 0.776629 & 0.780256 \\ \cline{2-11} 
\multirow{2}{*}{SVGAE} & Recall & \multicolumn{1}{r}{0.273485} & 0.263785 & 0.273485 & 0.294505 & 0.306035 & 0.315532 & 0.321633 & 0.32235 & 0.318589 \\
 & Precision & 0.494758 & 0.432536 & 0.494758 & 0.540298 & 0.576909 & 0.596037 & 0.603462 & 0.610673 & 0.604129 \\ \cline{2-11} 
\multirow{2}{*}{dynAE} & Recall & 0.243241 & 0.260874 & 0.249538 & 0.340724 & 0.363635 & 0.430242 & 0.027614 & 0.425404 & 0.424393 \\
 & Precision & 0.484355 & 0.519891 & 0.58471 & 0.675628 & 0.669945 & 0.667173 & 0.064773 & 0.764048 & 0.763721 \\ \cline{2-11} 
\multirow{2}{*}{dynAERNN} & Recall & \textbf{0.294509} & \textbf{0.358547} & \textbf{0.375264} & \textbf{0.409208} & \textbf{0.445241} & \textbf{0.481598} & \textbf{0.436302} & 0.432607 & 0.43122 \\
 & Precision & \textbf{0.501646} & \textbf{0.555114} & \textbf{0.670827} & \textbf{0.712004} & \textbf{0.786749} & \textbf{0.794443} & \textbf{0.776476} & 0.773307 & 0.771579 \\ \hline
 \multirow{2}{*}{\textbf{Dyn-VGAE1}} & Recall & 0.165353 & 0.253836 & 0.26011 & 0.280628 & 0.291986 & 0.301531 & 0.307532 & 0.31024 & 0.315377 \\
 & Precision & 0.30464 & 0.416175 & 0.47301 & 0.516912 & 0.551996 & 0.570365 & 0.57744 & 0.586087 & 0.588087 \\ \cline{2-11} 
\multirow{2}{*}{\textbf{Dyn-VGAE2}} & Recall & 0.16685 & 0.260054 & 0.269341 & 0.290558 & 0.302173 & 0.312019 & 0.317569 & 0.319883 & 0.325352 \\
 & Precision & 0.308355 & 0.426376 & 0.487754 & 0.53296 & 0.56928 & 0.588481 & 0.595397 & 0.604296 & 0.606698
\end{tabular}
\end{table*}

\begin{table*}[]
\caption{Precision and Recall for recommendation on Hep-th dataset}
\label{tab6}
\small
\begin{tabular}{l|l|lllllllll}
\multicolumn{1}{c|}{\textbf{Method}} & \multicolumn{1}{c|}{} & \multicolumn{1}{c}{\textbf{k=2}} & \multicolumn{1}{c}{\textbf{k=3}} & \multicolumn{1}{c}{\textbf{k=4}} & \multicolumn{1}{c}{\textbf{k=5}} & \multicolumn{1}{c}{\textbf{k=6}} & \multicolumn{1}{c}{\textbf{k=7}} & \multicolumn{1}{c}{\textbf{k=8}} & \multicolumn{1}{c}{\textbf{k=9}} & \multicolumn{1}{c}{\textbf{k=10}} \\ \hline
\multirow{2}{*}{node2vec} & Recall & \multicolumn{1}{r}{0.32482} & 0.465231 & 0.55624 & 0.585563 & 0.609622 & 0.646217 & 0.569039 & 0.598079 & 0.547933 \\
 & Precision & \multicolumn{1}{r}{0.44415} & 0.585353 & 0.67445 & 0.726131 & 0.739472 & 0.766933 & 0.68246 & 0.687628 & 0.630987 \\ \cline{2-11} 
\multirow{2}{*}{DeepWalk} & Recall & \multicolumn{1}{r}{0.315802} & 0.468277 & 0.550393 & 0.572584 & 0.607141 & 0.627954 & 0.568508 & 0.594742 & 0.545312 \\
 & Precision & \multicolumn{1}{r}{0.433653} & 0.586132 & 0.668169 & 0.709337 & 0.736379 & 0.745321 & 0.681347 & 0.683836 & 0.628118 \\ \cline{2-11} 
\multirow{2}{*}{SVGAE} & Recall & \multicolumn{1}{r}{0.378678} & 0.511046 & 0.580978 & 0.600192 & 0.614316 & 0.643828 & 0.567718 & 0.551842 & 0.549033 \\
 & Precision & 0.510663 & 0.643486 & 0.703221 & 0.720982 & 0.751181 & 0.763746 & 0.682238 & 0.685163 & 0.632336 \\ \cline{2-11} 
\multirow{2}{*}{dynAE} & Recall & 0.337487 & 0.059347 & 0.530609 & 0.57412 & 0.575352 & 0.558555 & 0.5437 & 0.537371 & 0.528373 \\
 & Precision & 0.452566 & 0.073831 & 0.644052 & 0.688662 & 0.685259 & 0.666084 & 0.649728 & 0.640798 & 0.601132 \\ \cline{2-11} 
\multirow{2}{*}{dynAERNN} & Recall & 0.373237 & 0.508547 & 0.562752 & 0.599208 & 0.605728 & 0.581598 & 0.565275 & 0.557373 & 0.544694 \\
 & Precision & 0.501881 & 0.635114 & 0.665775 & 0.722004 & 0.722058 & 0.694443 & 0.675759 & 0.664076 & 0.620038 \\ \hline
\multirow{2}{*}{\textbf{Dyn-VGAE1}} & Recall & 0.389606 & 0.522072 & 0.591764 & 0.602518 & 0.627004 & 0.645771 & 0.581532 & 0.608694 & 0.55852 \\
 & Precision & 0.521951 & 0.646657 & 0.714264 & 0.743664 & 0.76054 & 0.766375 & 0.697046 & 0.699577 & 0.642631 \\ \cline{2-11} 
\multirow{2}{*}{\textbf{Dyn-VGAE2}} & Recall & \textbf{0.39046} &\textbf{ 0.522811} & \textbf{0.59121} &\textbf{ 0.602739} & \textbf{0.626602} & \textbf{0.64676} & \textbf{0.581511} & \textbf{0.610082} & \textbf{0.560516} \\
 & Precision & \textbf{0.522826} & \textbf{0.647528} & \textbf{0.713929} & \textbf{0.743913} & \textbf{0.760167} & \textbf{0.767543} &\textbf{ 0.697636} & \textbf{0.701501} & \textbf{0.645338}
\end{tabular}
\end{table*}

\begin{table*}[]
\caption{Precision and Recall for recommendation on St-Ov dataset}
\label{tab7}
\small
\begin{tabular}{l|l|lllllllll}
\multicolumn{1}{c|}{\textbf{Method}} & \multicolumn{1}{c|}{} & \multicolumn{1}{c}{\textbf{k=2}} & \multicolumn{1}{c}{\textbf{k=3}} & \multicolumn{1}{c}{\textbf{k=4}} & \multicolumn{1}{c}{\textbf{k=5}} & \multicolumn{1}{c}{\textbf{k=6}} & \multicolumn{1}{c}{\textbf{k=7}} & \multicolumn{1}{c}{\textbf{k=8}} & \multicolumn{1}{c}{\textbf{k=9}} & \multicolumn{1}{c}{\textbf{k=10}} \\ \hline
\multirow{2}{*}{node2vec} & Recall & \multicolumn{1}{r}{0.199371} & 0.170521 & 0.151717 & 0.142989 & 0.128116 & 0.125163 & 0.120154 & 0.116243 & 0.113564 \\
 & Precision & \multicolumn{1}{r}{0.343357} & 0.271554 & 0.229346 & 0.207767 & 0.185266 & 0.17563 & 0.163032 & 0.154961 & 0.148131 \\ \cline{2-11} 
\multirow{2}{*}{DeepWalk} & Recall & \multicolumn{1}{r}{0.199072} & 0.170562 & 0.151378 & 0.142644 & 0.126195 & 0.122637 & 0.118336 & 0.119307 & 0.115129 \\
 & Precision & \multicolumn{1}{r}{0.342977} & 0.271374 & 0.228719 & 0.207191 & 0.182268 & 0.172017 & 0.161157 & 0.158638 & 0.150188 \\ \cline{2-11} 
\multirow{2}{*}{SVGAE} & Recall & \multicolumn{1}{r}{0.200822} & 0.170041 & 0.155608 & 0.144517 & 0.131738 & 0.128646 & 0.125685 & 0.122943 & 0.121674 \\
 & Precision & 0.34631 & 0.274437 & 0.236112 & 0.212186 & 0.191121 & 0.181621 & 0.171127 & 0.164683 & 0.15869 \\ \cline{2-11} 
\multirow{2}{*}{dynAE} & Recall & 0.088628 & 0.077637 & 0.067938 & 0.062643 & 0.05903 & 0.054914 & 0.054631 & 0.053159 & 0.049165 \\
 & Precision & 0.158084 & 0.125986 & 0.104936 & 0.093048 & 0.085407 & 0.077678 & 0.074203 & 0.071328 & 0.064948 \\ \cline{2-11} 
\multirow{2}{*}{dynAERNN} & Recall & 0.108056 & 0.094056 & 0.082202 & 0.077242 & 0.071827 & 0.067795 & 0.066443 & 0.065219 & 0.064853 \\
 & Precision & 0.191947 & 0.152271 & 0.127425 & 0.114656 & 0.104315 & 0.095892 & 0.09075 & 0.087338 & 0.084257 \\ \hline
\multirow{2}{*}{\textbf{Dyn-VGAE1}} & Recall & \textbf{0.204545} & \textbf{0.177749} & \textbf{0.160533} & \textbf{0.151073} & \textbf{0.13851} &\textbf{ 0.136031} & \textbf{0.134245} & \textbf{0.133499} &\textbf{ 0.135583} \\
 & Precision & \textbf{0.352988} & \textbf{0.284164 }& \textbf{0.244354} & \textbf{0.222065} & \textbf{0.202489} & \textbf{0.193704} & \textbf{0.185364 }& \textbf{0.181157} & \textbf{0.179375} \\ \cline{2-11} 
\multirow{2}{*}{\textbf{Dyn-VGAE2}} & Recall & 0.20165 & 0.173873 & 0.157185 & 0.147463 & 0.134296 & 0.128952 & 0.128332 & 0.126622 & 0.127372 \\
 & Precision & 0.347869 & 0.277905 & 0.238164 & 0.21595 & 0.194931 & 0.182753 & 0.174999 & 0.170079 & 0.166873
\end{tabular}
\end{table*}

\begin{table*}[]
\caption{Analysis of parameter $\gamma$ for recommendation}
\label{tab8}
\small
\begin{tabular}{c|l|l|lllllllll}
\textbf{Dataset} & \multicolumn{1}{c|}{\textbf{a}} & \multicolumn{1}{c|}{} & \multicolumn{1}{c}{\textbf{k=2}} & \multicolumn{1}{c}{\textbf{k=3}} & \multicolumn{1}{c}{\textbf{k=4}} & \multicolumn{1}{c}{\textbf{k=5}} & \multicolumn{1}{c}{\textbf{k=6}} & \multicolumn{1}{c}{\textbf{k=7}} & \multicolumn{1}{c}{\textbf{k=8}} & \multicolumn{1}{c}{\textbf{k=9}} & \multicolumn{1}{c}{\textbf{k=10}} \\ \hline
\multirow{6}{*}{\textbf{Hep-th}} & \multirow{2}{*}{0.6} & Recall & \multicolumn{1}{r}{0.389327} & 0.523287 & 0.594644 & 0.604486 & 0.628978 & 0.647625 & 0.582096 & 0.609569 & 0.561969 \\
 &  & Precision & 0.521504 & 0.648319 & 0.717725 & 0.746177 & 0.762935 & 0.768519 & 0.698145 & 0.701224 & 0.64713 \\ \cline{2-12} 
 & \multirow{2}{*}{1} & Recall & \multicolumn{1}{r}{0.388978} & 0.522966 & 0.592967 & 0.603635 & 0.629609 & 0.649588 & 0.58224 & 0.609715 & 0.558395 \\
 &  & Precision & 0.52145 & 0.648012 & 0.715753 & 0.745351 & 0.763854 & 0.770862 & 0.698341 & 0.701171 & 0.643245 \\ \cline{2-12} 
 & \multirow{2}{*}{1.2} & Recall & 0.389843 & 0.522133 & 0.592613 & 0.603883 & 0.628463 & 0.64613 & 0.581418 & 0.606744 & 0.559629 \\
 &  & Precision & 0.522121 & 0.647054 & 0.715439 & 0.745214 & 0.761996 & 0.766968 & 0.697532 & 0.698231 & 0.644725 \\ \hline
\multirow{6}{*}{\textbf{AS}} & \multirow{2}{*}{0.2} & Recall & \multicolumn{1}{r}{0.17086} & 0.266335 & 0.276734 & 0.297303 & 0.309268 & 0.317995 & 0.323751 & 0.325599 & 0.332499 \\
 &  & Precision & 0.314745 & 0.436361 & 0.499391 & 0.544163 & 0.581436 & 0.599712 & 0.607367 & 0.616013 & 0.620238 \\ \cline{2-12} 
 & \multirow{2}{*}{0.5} & Recall & \multicolumn{1}{r}{0.170643} & 0.264457 & 0.273346 & 0.293521 & 0.305441 & 0.314985 & 0.320376 & 0.322351 & 0.32855 \\
 &  & Precision & 0.313586 & 0.432619 & 0.493706 & 0.53772 & 0.574147 & 0.593054 & 0.59961 & 0.607748 & 0.611074 \\ \cline{2-12} 
 & \multirow{2}{*}{1.2} & Recall & \multicolumn{1}{r}{0.160112} & 0.247704 & 0.251588 & 0.270774 & 0.281948 & 0.290562 & 0.29635 & 0.298402 & 0.3036 \\
 &  & Precision & 0.296535 & 0.405931 & 0.458814 & 0.500643 & 0.53471 & 0.551478 & 0.558458 & 0.565753 & 0.56792 \\ \hline
\multicolumn{1}{l|}{\multirow{6}{*}{\textbf{St-Ov}}} & \multirow{2}{*}{0.5} & Recall & 0.20672 & 0.18231 & 0.167141 & 0.162219 & 0.14719 & 0.144816 & 0.142089 & 0.140077 & 0.14082 \\
\multicolumn{1}{l|}{} &  & Precision & 0.35675 & 0.291245 & 0.253335 & 0.236867 & 0.213672 & 0.20471 & 0.194035 & 0.188426 & 0.185214 \\ \cline{2-12} 
\multicolumn{1}{l|}{} & \multirow{2}{*}{0.7} & Recall & 0.203847 & 0.175776 & 0.158857 & 0.149699 & 0.135384 & 0.130333 & 0.129943 & 0.129038 & 0.129972 \\
\multicolumn{1}{l|}{} &  & Precision & 0.351087 & 0.281454 & 0.241279 & 0.219523 & 0.197625 & 0.186804 & 0.17955 & 0.174856 & 0.171854 \\ \cline{2-12} 
\multicolumn{1}{l|}{} & \multirow{2}{*}{1} & Recall & 0.204545 & 0.177749 & 0.160533 & 0.151073 & 0.13851 & 0.136031 & 0.134245 & 0.133499 & 0.135583 \\
\multicolumn{1}{l|}{} &  & Precision & 0.352988 & 0.284164 & 0.244354 & 0.222065 & 0.202489 & 0.193704 & 0.185364 & 0.181157 & 0.179375
\end{tabular}
\end{table*}

\subsection{Recommendation Task}
Recommendation is a challenging task, especially in dynamic graphs. A recommendation task aims to suggest potential relations to users in many networks such as coauthorship, communication, and interaction networks. For example, in the coauthorship network \textbf{Hep-th}, we recommend co-authors to researchers by learning their embeddings over time. In \cite{zuo2018embedding}, temporal recommendation is defined as recommending new connections for a node at time $t$ by using obtained embeddings from previous time points. Here, we use the learned embedding at time $t-1$ to rank nodes for recommending top-$k$ possible relations for the graph $G_t$. Our ranking score is based on the cosine similarity of embedding vectors of nodes. We use Precision@$k$ and Recall@$k$ as evaluation measures where the value $k$ varies from 2 to 10. The number of nodes is different for each $k$ because we select common nodes of consecutive times with more than $k$ neighbors. Tables \ref{tab5}, \ref{tab6}, and \ref{tab7} show the average Precision@$k$ and Recall@$k$ over time from $1$ to $T$ for three datasets: the coauthorship network \textbf{Hep-th}, communication network \textbf{AS}, and interaction user network \textbf{St-ov}. From the results, it can be seen that Dyn-VGAE performs better than other compared methods on \textbf{Hep-th} and \textbf{St-ov}; it obtains higher Precision@$k$ and Recall@$k$ for all different $k$ values. However, the performance of Dyn-VGAE decreases on \textbf{AS} and dynAERNN performs the best. Also, we can see that Dyn-VGA1 performs better than Dyn-VGA2 on the dataset \textbf{St-ov} due to characteristics of this dataset. As mentioned previously, the reason is that on this dataset considering smaller length $l$ is more suitable as we observed the same behaviour for the link predication task. It is worth mentioning that as the numbers of nodes for different $k$'s are different, we cannot see a decreasing trend by increasing $k$ for either Precision@$k$ or Recall@$k$.  

\begin{figure*}
    \centering
    \includegraphics[scale=0.7]{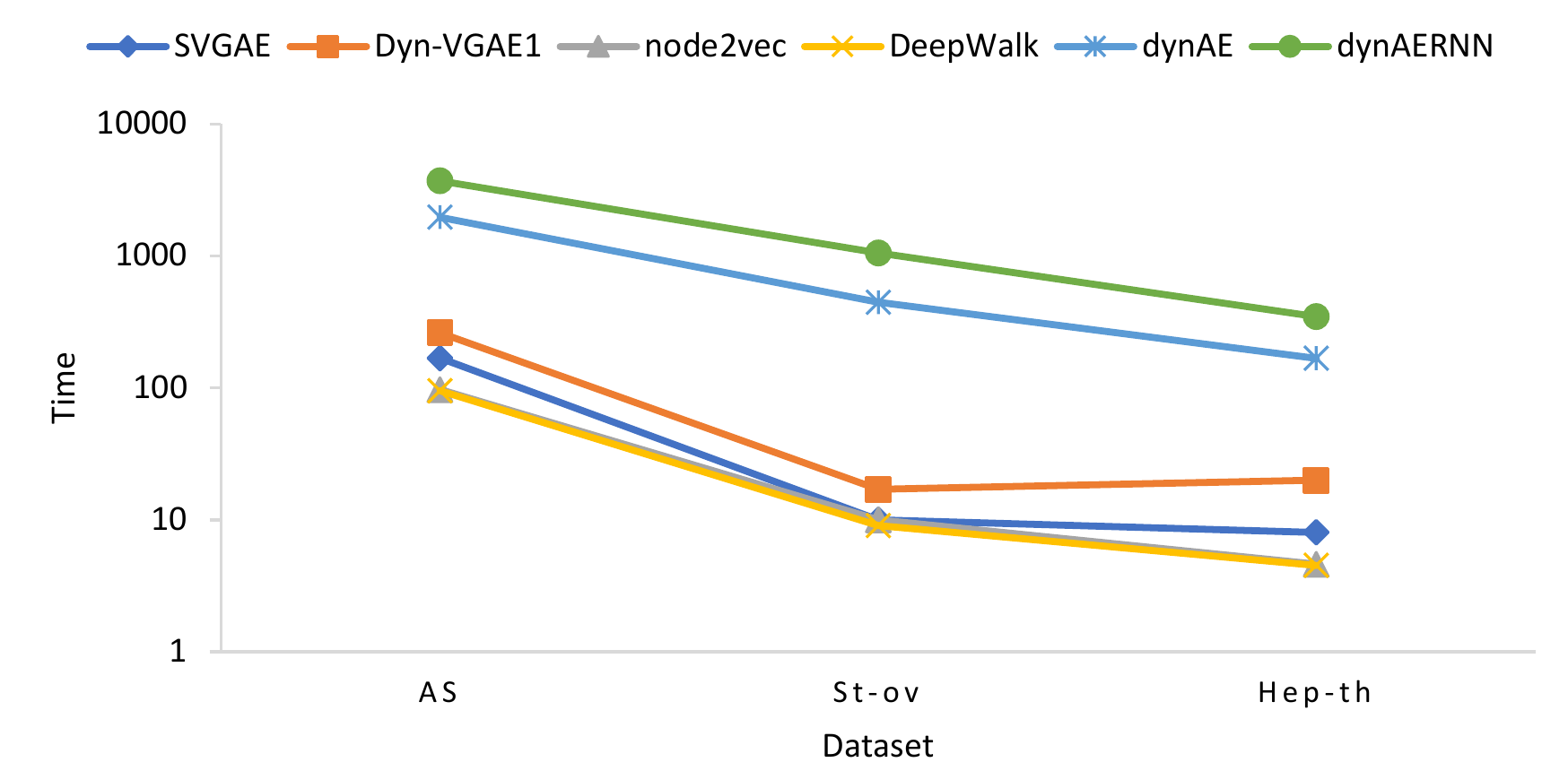}
    \caption{Computation time of embedding methods for the four timestamps on each dataset.}
    \label{fig2}
\end{figure*}
\subsection{The effect of temporal smoothness $\gamma$}
We study the effect of $\gamma$ on the performance of Dyn-VGAE. The parameter $\gamma$ can be fine-tuned to balance the weights of the local structure of the graph and the effect of previous graphs based on the data and the requirement of the task. We examine how the changes in $\gamma$ can affect the results. We vary $\alpha$ from 0 to 3. If $\gamma = 0$, the dynamic representations are only learning the local structure of the graph. By increasing the value of $\gamma$, we force the learned latent space of the graph to be aligned with the space of previous graphs.  Our experiments show that Dyn-VGAE has the best performance when $\gamma \in [0.1,2.5]$ and it starts decreasing for $\gamma > 2.5$. We report the results of our analysis for three values of $\gamma$ for each dataset in Tables \ref{tab3:title},\ref{tab4:title}, and \ref{tab8} for Dyn-VGAE with $l=1$. We observe the same behavior for Dyn-VGAE with $l=2$.
\subsection{Time complexity analysis}
We compare our method with all baseline models in terms of running time (in seconds) on three datasets AS, Hep-th, and  St-Ov at their first four time steps. All experiments are performed on a windows X-64 machine with 7 cores, 64 GB RAM and a clock speed of 3.6 GHz. From Figure \ref{fig2}, we observe that all static methods are faster than dynamic methods. The reason is that they only compute embedding vectors for each time step without adjusting these embedding vectors. Among compared dynamic embedding methods, Dyn-VGAE1 is the fastest. It is worth noting that the computation time of Dyn-VGAE1 is not significantly larger than SVGA while it outperforms SVGA in terms of the accuracy with a large margin. To further decrease the running time of Dyn-VGAE1, we are interested in developing a distributed version of our method by using proposed strategies in \cite{simonyan2014very,liu2018usability,dean2012large} as future work.


\section{Conclusions and Future Work}

In this paper, a dynamic joint autoencoder is proposed to embed a dynamic network into a low-dimensional latent space. For capturing evolving dependencies, we define a probabilistic smoothness term which changes the learning process of a graph variational autoencoder. The proposed joint framework provides a model where autoencoders can share their learned latent vectors across time stamps. The basic idea of the approach is the sharing of learned information of the current graph snapshot with previous graph snapshots for common nodes while each autoencoder works on its own specific graph. Dyn-VGAE simultaneously learns the latent representations of a dynamic network and aligns them across time. The experimental results show that Dyn-VGAE can significantly outperform the state-of-the-art methods on the node classification, link predication, and recommendation tasks. In the future, we are interested in applying different variants of joint deep learning architectures to extract the dynamic latent space of a dynamic network and developing a distributed model to increase the speed of training. Also, we will investigate a joint approach for other deep graph embedding methods.

\bibliographystyle{splncs04}
\bibliography{ref}

\begin{thebibliography}{10}
\providecommand{\url}[1]{\texttt{#1}}
\providecommand{\urlprefix}{URL }
\providecommand{\doi}[1]{https://doi.org/#1}

\bibitem{cai2018comprehensive}
Cai, H., Zheng, V.W., Chang, K.C.C.: A comprehensive survey of graph embedding:
  Problems, techniques, and applications. IEEE Transactions on Knowledge and
  Data Engineering  \textbf{30}(9),  1616--1637 (2018)

\bibitem{cao2015grarep}
Cao, S., Lu, W., Xu, Q.: Grarep: Learning graph representations with global
  structural information. In: Proceedings of the 24th ACM international on
  conference on information and knowledge management. pp. 891--900. ACM (2015)

\bibitem{cao2016deep}
Cao, S., Lu, W., Xu, Q.: Deep neural networks for learning graph
  representations. In: Thirtieth AAAI Conference on Artificial Intelligence
  (2016)

\bibitem{cui2018survey}
Cui, P., Wang, X., Pei, J., Zhu, W.: A survey on network embedding. IEEE
  Transactions on Knowledge and Data Engineering  (2018)

\bibitem{dean2012large}
Dean, J., Corrado, G., Monga, R., Chen, K., Devin, M., Mao, M., Senior, A.,
  Tucker, P., Yang, K., Le, Q.V., et~al.: Large scale distributed deep
  networks. In: Advances in neural information processing systems. pp.
  1223--1231 (2012)

\bibitem{epstein2018joint}
Epstein, B., Meir, R., Michaeli, T.: Joint autoencoders: a flexible
  meta-learning framework. In: Joint European Conference on Machine Learning
  and Knowledge Discovery in Databases. pp. 494--509. Springer (2018)

\bibitem{ghasedi2017deep}
Ghasedi~Dizaji, K., Herandi, A., Deng, C., Cai, W., Huang, H.: Deep clustering
  via joint convolutional autoencoder embedding and relative entropy
  minimization. In: Proceedings of the ICCV. pp. 5736--5745 (2017)

\bibitem{goyal2018graph}
Goyal, P., Ferrara, E.: Graph embedding techniques, applications, and
  performance: A survey. Knowledge-Based Systems  \textbf{151},  78--94 (2018)

\bibitem{goyal2018capturing}
Goyal, P., Hosseinmardi, H., Ferrara, E., Galstyan, A.: Capturing edge
  attributes via network embedding. IEEE Transactions on Computational Social
  Systems  \textbf{5}(4),  907--917 (2018)

\bibitem{goyal2018dyngem}
Goyal, P., Kamra, N., He, X., Liu, Y.: Dyngem: Deep embedding method for
  dynamic graphs. In: IJCAI International Workshop on Representation Learning
  for Graphs (2017)

\bibitem{goyal2018dynamicgem}
Goyal, P., Rokka~Chhetri, S., Mehrabi, N., Ferrara, E., Canedo, A.: Dynamicgem:
  A library for dynamic graph embedding methods. arXiv preprint
  arXiv:1811.10734  (2018)

\bibitem{grover2016node2vec}
Grover, A., Leskovec, J.: node2vec: Scalable feature learning for networks. In:
  Proceedings of the 22nd ACM SIGKDD. pp. 855--864. ACM (2016)

\bibitem{hamilton2017inductive}
Hamilton, W., Ying, Z., Leskovec, J.: Inductive representation learning on
  large graphs. In: Advances in Neural Information Processing Systems. pp.
  1024--1034 (2017)

\bibitem{hamilton2017representation}
Hamilton, W.L., Ying, R., Leskovec, J.: Representation learning on graphs:
  Methods and applications. arXiv preprint arXiv:1709.05584  (2017)

\bibitem{huang2018multimodal}
Huang, F., Zhang, X., Li, C., Li, Z., He, Y., Zhao, Z.: Multimodal network
  embedding via attention based multi-view variational autoencoder. In:
  Proceedings of the 2018 ACM on International Conference on Multimedia
  Retrieval. pp. 108--116. ACM (2018)

\bibitem{kingma2013auto}
Kingma, D.P., Welling, M.: Auto-encoding variational bayes. arXiv preprint
  arXiv:1312.6114  (2013)

\bibitem{kipf2016semi}
Kipf, T.N., Welling, M.: Semi-supervised classification with graph
  convolutional networks. arXiv preprint arXiv:1609.02907  (2016)

\bibitem{kipf2016variational}
Kipf, T.N., Welling, M.: Variational graph auto-encoders. stat  \textbf{1050},
  ~21 (2016)

\bibitem{kumar2018learning}
Kumar, S., Zhang, X., Leskovec, J.: Learning dynamic embeddings from temporal
  interactions. arXiv preprint arXiv:1812.02289  (2018)

\bibitem{leskovec2015snap}
Leskovec, J., Krevl, A.: $\{$SNAP Datasets$\}$:$\{$Stanford$\}$ large network
  dataset collection  (2015)

\bibitem{liu2018usability}
Liu, J., Dutta, J., Li, N., Kurup, U., Shah, M.: Usability study of distributed
  deep learning frameworks for convolutional neural networks  (2018)

\bibitem{mahdavi2018dynnode2vec}
Mahdavi, S., Khoshraftar, S., An, A.: dynnode2vec: Scalable dynamic network
  embedding. In: 2018 IEEE Big Data. pp. 3762--3765. IEEE (2018)

\bibitem{nguyen2018continuous}
Nguyen, G.H., Lee, J.B., Rossi, R.A., Ahmed, N.K., Koh, E., Kim, S.:
  Continuous-time dynamic network embeddings. In: Companion of the The Web
  Conference 2018 on The Web Conference 2018. pp. 969--976. International World
  Wide Web Conferences Steering Committee (2018)

\bibitem{ou2016asymmetric}
Ou, M., Cui, P., Pei, J., Zhang, Z., Zhu, W.: Asymmetric transitivity
  preserving graph embedding. In: Proceedings of the 22nd ACM SIGKDD
  international conference on Knowledge discovery and data mining. pp.
  1105--1114. ACM (2016)

\bibitem{pan2018adversarially}
Pan, S., Hu, R., Long, G., Jiang, J., Yao, L., Zhang, C.: Adversarially
  regularized graph autoencoder for graph embedding. In: Proceedings of the
  27th IJCAI. pp. 2609--2615. AAAI Press (2018)

\bibitem{pan2016jointly}
Pan, Y., Mei, T., Yao, T., Li, H., Rui, Y.: Jointly modeling embedding and
  translation to bridge video and language. In: Proceedings of the IEEE
  conference on computer vision and pattern recognition. pp. 4594--4602 (2016)

\bibitem{perozzi2014deepwalk}
Perozzi, B., Al-Rfou, R., Skiena, S.: Deepwalk: Online learning of social
  representations. In: Proceedings of the 20th ACM SIGKDD. pp. 701--710. ACM
  (2014)

\bibitem{ren2016joint}
Ren, Z., Jin, H., Lin, Z., Fang, C., Yuille, A.: Joint image-text
  representation by gaussian visual-semantic embedding. In: Proceedings of the
  24th ACM international conference on Multimedia. pp. 207--211. ACM (2016)

\bibitem{sarkar2006dynamic}
Sarkar, P., Moore, A.W.: Dynamic social network analysis using latent space
  models. In: Advances in Neural Information Processing Systems. pp. 1145--1152
  (2006)

\bibitem{simonyan2014very}
Simonyan, K., Zisserman, A.: Very deep convolutional networks for large-scale
  image recognition. arXiv preprint arXiv:1409.1556  (2014)

\bibitem{tang2015line}
Tang, J., Qu, M., Wang, M., Zhang, M., Yan, J., Mei, Q.: Line: Large-scale
  information network embedding. In: Proceedings of the 24th WWW. pp.
  1067--1077. International World Wide Web Conferences Steering Committee
  (2015)

\bibitem{tang2008extraction}
Tang, J., Zhang, J., Yao, L., Li, J.: Extraction and mining of an academic
  social network. In: Proceedings of the 17th WWW. pp. 1193--1194. ACM (2008)

\bibitem{tang2008arnetminer}
Tang, J., Zhang, J., Yao, L., Li, J., Zhang, L., Su, Z.: Arnetminer: extraction
  and mining of academic social networks. In: Proceedings of the 14th ACM
  SIGKDD. pp. 990--998. ACM (2008)

\bibitem{trivedi2018representation}
Trivedi, R., Farajtbar, M., Biswal, P., Zha, H.: Representation learning over
  dynamic graphs. arXiv preprint arXiv:1803.04051  (2018)

\bibitem{tu2016max}
Tu, C., Zhang, W., Liu, Z., Sun, M., et~al.: Max-margin deepwalk:
  Discriminative learning of network representation. In: IJCAI. pp. 3889--3895
  (2016)

\bibitem{wang2016structural}
Wang, D., Cui, P., Zhu, W.: Structural deep network embedding. In: Proceedings
  of the 22nd ACM SIGKDD. pp. 1225--1234. ACM (2016)

\bibitem{wang2017community}
Wang, X., Cui, P., Wang, J., Pei, J., Zhu, W., Yang, S.: Community preserving
  network embedding. In: Thirty-First AAAI Conference on Artificial
  Intelligence (2017)

\bibitem{xu2017embedding}
Xu, L., Wei, X., Cao, J., Yu, P.S.: Embedding of embedding (eoe): Joint
  embedding for coupled heterogeneous networks. In: Proceedings of the Tenth
  ACM International Conference on Web Search and Data Mining. pp. 741--749. ACM
  (2017)

\bibitem{yu2018netwalk}
Yu, W., Cheng, W., Aggarwal, C.C., Zhang, K., Chen, H., Wang, W.: Netwalk: A
  flexible deep embedding approach for anomaly detection in dynamic networks.
  In: Proceedings of the 24th ACM SIGKDD. pp. 2672--2681. ACM (2018)

\bibitem{zhou2018dynamic}
Zhou, L., Yang, Y., Ren, X., Wu, F., Zhuang, Y.: Dynamic network embedding by
  modeling triadic closure process. In: Thirty-Second AAAI (2018)

\bibitem{zhuscalable}
Zhu, L., Ver~Steeg, G., Galstyan, A.: Scalable link prediction in dynamic
  networks via non-negative matrix factorization  (2018)

\bibitem{zuo2018embedding}
Zuo, Y., Liu, G., Lin, H., Guo, J., Hu, X., Wu, J.: Embedding temporal network
  via neighborhood formation. In: Proceedings of the 24th ACM SIGKDD. pp.
  2857--2866. ACM (2018)

\end{thebibliography}
\end{document}